\documentclass[runningheads]{llncs}

\usepackage[T1]{fontenc}
\usepackage{graphicx}
\usepackage{booktabs}
\usepackage{amsmath,amsfonts}
\usepackage{cite}
\usepackage{hyperref}
\usepackage{xcolor}

\begin{document}

\title{LongAgent: History-Guided Agentic Search for Longitudinal Outcome Prediction}
\titlerunning{History-Guided Agentic Search for Longitudinal Outcome Prediction}

\author{
Siyao Wang\inst{1,2}\thanks{Corresponding author.} \and
Florian Guitton\inst{1} \and
Shuojie Fu\inst{1} \and
Guanyu Tao\inst{1} \and
Kai Sun\inst{1} \and
Wenjia Bai\inst{1,2,3}
}
\authorrunning{S. Wang et al.}
\institute{
Data Science Institute, Imperial College London, London, UK
\and
Department of Computing, Imperial College London, London, UK
\and
Department of Brain Sciences, Imperial College London, London, UK\\
\email{s.wang18@imperial.ac.uk}
}

\maketitle

\begin{abstract}
Extracting informative representations from longitudinal data that can predict future outcomes remains a critical challenge in medicine. Medical datasets are inherently heterogeneous, consisting of a large number of variables collected from different sources, sampled with different temporal spacings, and representing different aspects of human health status. This requires identifying those variables with predictive value, processing longitudinal information, and integrating multiple variables for outcome prediction. Here, we propose a novel agent-based approach, LongAgent, that can autonomously search over combinations of variable sets, temporal windows and longitudinal aggregation functions, and identify candidates with promising predictive performance. LongAgent utilises a history memory of previous searches and numerical evidence to guide subsequent exploration. On synthetic data, LongAgent achieves a mean prediction RMSE of 1.7376 and improves over the strongest non-agent baseline by 0.0151 (95\% CI: $[0.0045,0.0260]$; $p=0.0273$). On a real clinical dataset, it performs comparably to the best baseline.

\keywords{Agentic search \and Predictive modelling \and Longitudinal data \and Longitudinal outcome prediction}
\end{abstract}

\section{Introduction}
In medical research, longitudinal outcome prediction involves collecting multimodal measurements of an individual, monitoring their longitudinal changes, and developing a predictive model that can learn multimodal features from the data for predicting future outcomes~\cite{cascarano2023longitudinal}. Typically, features, also known as predictors, are extracted by selecting a subset of recorded variables, followed by choosing an empirical time window and an aggregation function that can be applied to longitudinal measurements. Given the complexity of clinical longitudinal datasets and the exploratory nature of medical research, feature extraction is often performed in a hand-crafted manner and is time-consuming.

Repeated monitoring of outcomes adds another layer of complexity. In longitudinal studies, the outcome may be recorded at multiple time points for each participant. Predictors must therefore be constructed only from measurements preceding each observed outcome to avoid information leakage~\cite{cascarano2023longitudinal}. However, the appropriate duration of this preceding time window is often unknown. Dunn et al. found that the performance of predictive models using wearable device measurements varied with the monitoring time window and its proximity to the outcome date \cite{dunn2021wearable}. FIDDLE and flexible-window electronic health record (EHR) methods explore different ways to define the time window~\cite{tang2020fiddle,gupta2022flexiblewindow}. Based on the time window, different aggregation functions have been explored to integrate longitudinal measurements to form predictive features~\cite{fulcher2017hctsa,christ2018tsfresh,lubba2019catch22}.

Recently, AI agents, implemented as large language models (LLMs), have demonstrated great potential to either perform predefined workflows or discover new knowledge~\cite{collaco2026role, wang20263dmedagent, lu2026towards}. For longitudinal medical data analysis, PHIA has been proposed, which combines code generation and information retrieval to answer questions about wearable device data~\cite{merrill2026phia}. FeatEHR-LLM utilises an agent to generate executable feature extraction code for analysing irregular time series EHR data \cite{karami2026featehr}. CoDaS coordinates hypothesis generation, analysis, grounding, and validation for discovering wearable device biomarkers \cite{kim2026codas}. LLM-FE proposes validation-guided evolution to search over a space of feature transformation programs to process tabular data~\cite{abhyankar2026llmfe}.

These studies demonstrate the potential of LLM agents but primarily focus on generating code for feature transformation and data analysis. The proposed method, LongAgent, focuses on search for longitudinal outcome predictors. It uses accumulated evidence to guide the search for a combination of variable subset, time window and aggregation function that leads to predictive features for future outcomes. Our main contributions are:
\begin{enumerate}
\item We formulate the discovery of predictive features as a budgeted search problem, over a search space of variable subsets, time windows, and aggregation functions.
\item We introduce a history-guided search agent that combines search status, candidate history, coverage summary, and transition evidence, and uses a two-stage decision process to decide the next move.
\item We evaluate LongAgent against three non-agent search policies on a large number of synthetic datasets constructed with diverse settings and a real longitudinal cohort, and demonstrate LongAgent outperforms or on par with other search policies.
\end{enumerate}

\section{Methods}
\noindent\textbf{Problem formulation.}\label{sec:problem_formulation}
Consider a longitudinal medical dataset containing repeated measurements and outcomes from a cohort of participants. The objective is to extract predictive features from these measurements and fit a model onto the features to predict each participant's outcome. 

Denote the number of subjects as $N$, the number of recorded variables as $M$, and the dataset of all variables as $X=\{x_{i,j} | 1 \leq i \leq N, 1 \leq j  \leq M\}$, where $i$ denotes the subject index and $j$ denotes the variable index. Each variable $x_{i,j}$ is measured multiple times in the longitudinal study with $x_{i,j}[t]$ denoting the measurement at time $t$. Note that the time interval between measurements can be irregular for different subjects and for different variables. For each subject, the outcome is monitored longitudinally. Denote the set of outcomes as $Y=\{y_i\mid 1\leq i\leq N\}$, where $y_i$ denotes a sequence of outcomes for subject $i$, and $y_i[\tau]\in\mathbb{R}$ denotes the outcome observed at time $\tau$. 

The proposed AI agent searches over a space of candidates, defined by a variable subset, an outcome-relative time window, and an aggregation function. Each candidate $c$ is formulated as $c=(S,w,f)$, where $S\subseteq\{1,\ldots,M\}$ denotes the indices of the selected variables, $w$ denotes the time window preceding an outcome, and $f$ denotes an aggregation function applied to measurements within that window. Applying candidate $c$ to the longitudinal measurements of subject $i$ yields a feature vector or predictor $z_i$ for the selected variables $S$:
\begin{equation}
\begin{split}
z_{i,j} & = f\!\left(\{x_{i,j}[t]\mid \tau-w<t\leq\tau\}\right),
\qquad j\in S,\\
z_i & = \left(z_{i,j}\right)_{j\in S}
\end{split}
\end{equation}
Let $Z=\{z_i\mid 1\leq i\leq N\}$ denote the set of features or predictors. The performance of candidate $c$ can be evaluated by fitting a model onto $\{Z,Y\}$ and computing the prediction error $l(c)$. The objective is to find the optimal candidate $c$ that minimises $l(c)$.
\\

\begin{figure}[!t]
\centering
\includegraphics[width=\textwidth]{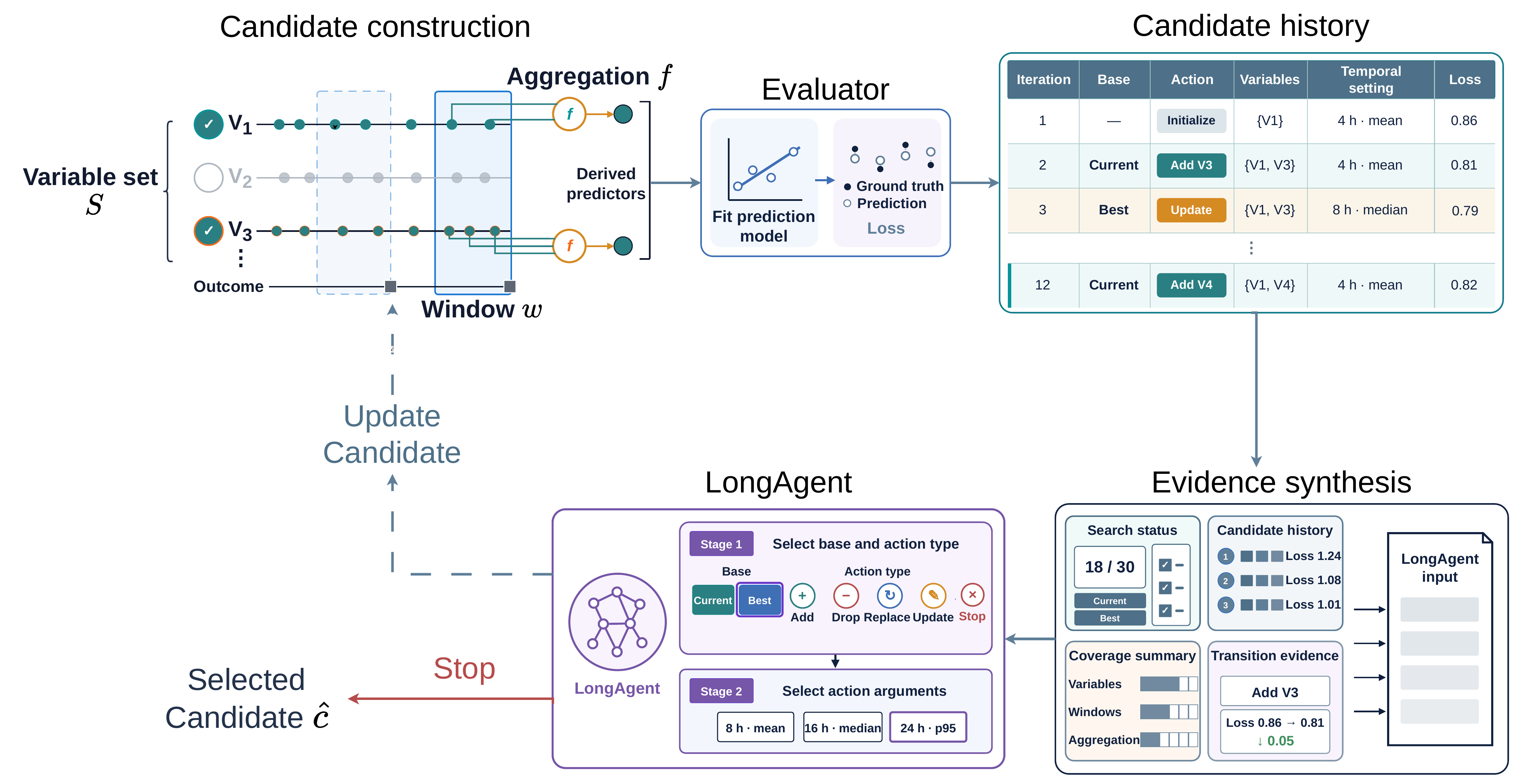}
\caption{History-guided agentic search. A candidate $c=(S,w,f)$ specifies a variable subset, an outcome-relative window, and an aggregation function. The evaluator fits a prediction model onto features extracted using $c$, and records the loss in the search history. The agent integrates four sources of evidence: search status, candidate history, coverage summary, and transition evidence, and then selects a base candidate and the next move. The agent either continues the search or stops when the budget is exhausted.}
\label{fig:agent_loop}
\end{figure}

\noindent\textbf{Iterative search.} Figure~\ref{fig:agent_loop} illustrates the procedure. The agent starts the search from an initial candidate $c^{(0)} = (S^{(0)}, w^{(0)}, f^{(0)})$, where $w^{(0)}$ and $f^{(0)}$ are predefined initial choices for the time window and aggregation function. Applying $w^{(0)}$ and $f^{(0)}$ to each recorded variable constructs one feature per variable. The variables are ranked by the absolute Pearson correlation between these features and the outcome, and $S^{(0)}$ is initialised with the highest-ranked variable.

After initialisation, the agent performs iterative search under an evaluation budget $B$, i.e. the maximum number of search steps. At iteration $n$, it sets either the current candidate or the previously best performing candidate to be the base candidate
$b^{(n)}=(S_b^{(n)},w_b^{(n)},f_b^{(n)})$, and proposes an action
$a^{(n)}\in\{\operatorname{add},\allowbreak
\operatorname{drop},\allowbreak
\operatorname{replace},\allowbreak
\operatorname{update},\allowbreak
\operatorname{stop}\}$ to modify the base candidate. For the non-stop actions, applying $a^{(n)}$ to the base candidate $b^{(n)}$ produces the
next candidate:
\begin{equation}
c^{(n)} =
\begin{cases}
(S_b^{(n)}\cup\{k\},w_b^{(n)},f_b^{(n)}),
& a^{(n)}=\operatorname{add}(k),\\
(S_b^{(n)}\setminus\{k\},w_b^{(n)},f_b^{(n)}),
& a^{(n)}=\operatorname{drop}(k),\\
((S_b^{(n)}\setminus\{k\})\cup\{k'\},w_b^{(n)},f_b^{(n)}),
& a^{(n)}=\operatorname{replace}(k,k'),\\
(S_b^{(n)},w',f'),
& a^{(n)}=\operatorname{update}(w',f').
\end{cases}
\end{equation}

The first three actions add, drop, or replace a recorded variable in $S$. The update action changes the time window $w$, the aggregation function $f$, or both. Previously evaluated candidates are excluded from the space of search, ensuring that $c^{(n)}$ has not been evaluated before. The stop action terminates the search without generating the next candidate.
\\

\noindent\textbf{History-guided search agent.} At iteration $n$, the agent receives history information $p^{(n)}$ coming from four sources $p^{(n)}=[s^{(n)},h^{(n-1)},g^{(n-1)},e^{(n-1)}]$, where:
\begin{enumerate}
\item Search status and constraints $s^{(n)}$: the current candidate, remaining budget, permitted variables, windows and aggregation functions, and legal candidate updates. It also includes the variable ranking used in initialisation.
\item Candidate history $h^{(n-1)}$: an ordered sequence $\{c^{(m)},l(c^{(m)})\}_{m=0}^{n-1}$ of previously evaluated candidates and their prediction errors.
\item Coverage summary $g^{(n-1)}$: the variables, window and aggregation combinations that have already been evaluated, which helps the agent define the unexplored space.
\item Transition evidence $e^{(n-1)}$: the previous transitions from base candidates to modified candidates, along with the change of prediction performance as evidence.
\end{enumerate}

LongAgent is implemented as an LLM, queried in two stages using different prompt templates. 
In the first stage, LongAgent selects a base candidate $b^{(n)}$ and an action type $a^{(n)}$ based on the evidence $p^{(n)}$. In the second stage, LongAgent is provided with the same evidence, including $p^{(n)}$, and the selected pair $(b^{(n)},a^{(n)})$ from the first stage and a list of parameters for action $a^{(n)}$. It then selects the parameter for $a^{(n)}$ from the list.
The selected action, equipped with the selected parameter, is applied to modify $b^{(n)}$ and generate the next candidate $c^{(n)}$, which is evaluated and added to the search history. When the LLM selects the stop action or the evaluation budget is exhausted, the method returns $\hat{c}=\arg\min_{c\in\mathcal E_B}l(c)$, where $\mathcal E_B$ denotes the set of all candidates evaluated within budget $B$.

\section{Experiments}
\noindent\textbf{Synthetic longitudinal datasets.} Synthetic data are generated in two stages: longitudinal measurements followed by outcome generation.

\noindent\textit{Stage 1: Longitudinal measurement generation.} For subject $i$ and variable $j$, the measurement at time $t$ is simulated as:
\begin{equation}
x_{i,j}[t]= \lambda_j \cdot u_i + d_{i,j} \cdot t
+ c_{i, j}\cdot \sin\!\left(\frac{2\pi h(t)}{24}\right)
+\epsilon_{i,j}[t].
\end{equation}
Here, $u_i$ denotes a subject-level effect shared across variables, $\lambda_j$ denotes the variable-level loading, $d_{i,j} \cdot t$ models linear temporal drift at time $t$, and $c_{i,j}\cdot \sin(2\pi h(t)/24)$ models the daily cycle with amplitude $c_{i,j}$. $h(t)$ denotes the hour of day, and $\epsilon_{i,j}[t]\sim\mathcal N(0,\sigma_X^2)$ denotes Gaussian measurement noise.
\\

\noindent\textit{Stage 2: Outcome generation.}
For subject $i$ and variable $j$, let $r_{i,j}[\tau]$ denote the aggregated measurement within the outcome-relative window: 
\begin{equation}
r_{i,j}[\tau]
=f^\star\!\left(\{x_{i,j}[t]\mid \tau-w^\star<t\leq\tau\}\right).
\end{equation}
Here, $w^\star$ denotes the temporal window and $f^\star$ denotes the aggregation function. The outcome $y_i^{(s)}[\tau]$ is generated from the combined effects of $r_{i,j}[\tau]$, $j=1,\ldots,M$, as follows:
\begin{equation}
\label{eq:outcome}
\begin{split}
y_i^{(s)}[\tau] = & \sum_{j\in S_{\mathrm{lin}}^{(s)}} \beta_j \cdot r_{i,j}[\tau]
+ \sum_{j\in S_{\mathrm{non}}^{(s)}} \gamma_j \cdot \lvert r_{i,j}[\tau]\rvert
+ \sum_{(j,k)\in\mathcal P_{\mathrm{int}}^{(s)}} \theta_{jk} \cdot r_{i,j}[\tau] \cdot r_{i,k}[\tau] \\
& + \delta_u^{(s)} \cdot u_i + \delta_b^{(s)} \cdot b_i + \xi_i^{(s)}[\tau].
\end{split}
\end{equation}
Here, the superscript $s\in\{1,\ldots,6\}$ identifies one of the six outcome generation settings (Table~\ref{tab:synthetic_settings}), each defined by a different combination of terms. $S_{\mathrm{lin}}^{(s)}$ and $S_{\mathrm{non}}^{(s)}$ denote the sets of variables with linear and absolute value effects, with coefficients $\beta_j$ and $\gamma_j$. $\mathcal P_{\mathrm{int}}^{(s)}$ denotes the set of interacting variable pairs, and $\theta_{jk}$ denotes the corresponding interaction coefficient for pair $(j,k)$. $u_i$ and $b_i$ denote shared and independent subject-level effects, and $\delta_u^{(s)}$ and $\delta_b^{(s)}$ denote their coefficients. Finally, $\xi_i^{(s)}[\tau]\sim\mathcal N(0,\sigma_Y^2)$ denotes the Gaussian outcome noise.
\\

\noindent\textbf{Clinical longitudinal dataset.} We use digital home monitoring data from IDEA-FAST, a prospective multinational study that followed over 1,000 participants for up to 24 weeks \cite{maetzler2026ideafast}. All data were pseudonymised and collected with ethical approval and written consent. Longitudinal measurements were recorded using a chest-worn VitalPatch and a lower-back Axivity AX6 inertial sensor, with 62 variables in total, consisting of 44 heart-rate variability, 4 VitalPatch summary, 7 activity, and 7 sit-to-stand or stand-to-sit variables. In terms of outcomes, 32,128 fatigue scores were recorded for 1,084 participants, which were normalised to z-scores.
\\

\noindent\textbf{Implementation.}
Each synthetic dataset contains 500 subjects, 100 variables measured every 30 minutes for 15 days, and one daily outcome per subject. We independently sample $u_i$ and $b_i$ from $\mathcal N(0,1)$, and $\lambda_j$, $d_{i,j}$, $c_{i,j}$, and $\epsilon_{i,j}[t]$ from $\mathcal N(0,1)$. The magnitude of the active coefficient is set to
$\kappa\in\{1,0.5,0.25\}$, and $|\beta_j|=|\gamma_j|=|\theta_{jk}|=\kappa$.
For each term, the coefficient is assigned randomly to either $+\kappa$ or $-\kappa$. $\sigma_X$ is set to 1 and $\sigma_Y \in \{0.5, 1\}$. We consider six outcome generation settings, 3 coefficient magnitudes for $\kappa$ and 2 outcome noise levels for $\sigma_Y$. For each combination, we generate 10 datasets using different random seeds, yielding 360 synthetic datasets in total.

\begin{table}[t]
\caption{Six synthetic outcome generation settings, using different combinations of linear terms, absolute value terms, interaction pairs, and subject-level effect coefficients in Eq.~\ref{eq:outcome}. $|\cdot|$ denotes the cardinality of a set, i.e. the number of variables.}
\label{tab:synthetic_settings}
\centering
\small
\begin{tabular}{lccccc}
\toprule
Setting & $|S_{\mathrm{lin}}^{(s)}|$ & $|S_{\mathrm{non}}^{(s)}|$
& $|\mathcal P_{\mathrm{int}}^{(s)}|$
& $\delta_u^{(s)}$ & $\delta_b^{(s)}$ \\
\midrule
Linear             & 6 & 0 & 0 & 0 & 0 \\
Nonlinear          & 2 & 4 & 0 & 0 & 0 \\
Interaction        & 2 & 0 & 2 & 0 & 0 \\
Shared factor      & 6 & 0 & 0 & $\kappa$ & 0 \\
Independent subject & 6 & 0 & 0 & 0 & $\kappa$ \\
Combined           & 2 & 2 & 1 & 0 & $\kappa$ \\
\bottomrule
\end{tabular}
\end{table}

For synthetic data, the search space of time windows is $\{8,\allowbreak 12,\allowbreak 16,\allowbreak 20,\allowbreak
24,\allowbreak 32,\allowbreak 48\}$ hours and the space of aggregation functions is $\{\mathrm{mean},\mathrm{median},\mathrm{p}_{95},\mathrm{p}_{05}\}$, $\mathrm{p}$ denoting percentile. For IDEA-FAST, the space of windows is $\{4,\allowbreak 8,\allowbreak 12,\allowbreak 16,\allowbreak
20,\allowbreak 24,\allowbreak 32,\allowbreak 48\}$ hours and the same space of aggregation functions is used. Synthetic and IDEA-FAST searches are initialised with $(w^{(0)},f^{(0)})=(12\text{h},\mathrm{mean})$ and $(8\text{h},\mathrm{mean})$, respectively. In both experiments, candidates contain at most ten variables, with a budget of $B = 30$ evaluations. Synthetic experiments use DeepSeek-V4-Flash \cite{deepseek2026v4} at temperature 0.01 with a 1,024-token limit and thinking disabled. For privacy reasons, IDEA-FAST experiments use gpt-oss-120b \cite{openai2025gptoss} at temperature 0.01.
\\

\noindent\textbf{Evaluation protocol and baseline methods.}
For both synthetic and real datasets, data is split subject-wise into 80\% for training and 20\% for test. The loss $l(c)$ is computed using five-fold cross-validation on the training set. The evaluator fits an elastic net model ($\alpha=0.1$ and equal $\ell_1$ and $\ell_2$ mixing, with $\ell_1$ ratio $=0.5$) and assesses the prediction performance in terms of the root mean squared error (RMSE). After the final candidate $\hat{c}$ is selected, the evaluator fits the model to the full training set and evaluates the prediction performance on the test set, in terms of the RMSE, mean absolute error (MAE), and the coefficient of determination ($R^2$). The prediction target is the simulated outcome for synthetic datasets, and fatigue z-scores for IDEA-FAST.

We compare LongAgent with three baselines. \textit{Greedy variable selection} retains fixed time window and aggregation function, and adds variables in decreasing order of absolute Pearson correlation~\cite{guyon2003feature,saeys2007feature}. \textit{Random candidate search} randomly samples a legal, previously unevaluated one-step neighbour of the current candidate. \textit{Rule-based candidate search} follows the correlation ranking before exploring time window and aggregation function that reduce validation loss of the evaluator. Confidence intervals are computed by resampling the ten seed blocks. Statistical comparison of two methods is performed using the two-sided Wilcoxon test on seed-level mean RMSE differences.

\section{Results}
\begin{table}[!ht]
\caption{Comparison of LongAgent with baseline methods on 360 synthetic datasets. Values are mean $\pm$ standard deviation across datasets; $p$-values compare RMSE between LongAgent and each baseline.}
\label{tab:main_results}
\centering
\footnotesize
\begin{tabular*}{\textwidth}{@{\extracolsep{\fill}}lcccc@{}}
\toprule
Method & RMSE $\downarrow$ & MAE $\downarrow$ & $R^2$ $\uparrow$ & $p$ \\
\midrule
Greedy variable selection & $1.7871 \pm 1.0060$ & $1.3357 \pm 0.6857$ & $0.3163 \pm 0.1756$ & 0.0020 \\
Random candidate search & $1.8094 \pm 1.0139$ & $1.3556 \pm 0.6957$ & $0.2999 \pm 0.1743$ & 0.0020 \\
Rule-based search & $1.7527 \pm 1.0024$ & $1.3091 \pm 0.6817$ & $0.3364 \pm 0.1916$ & 0.0273 \\
\textbf{LongAgent} & $\mathbf{1.7376 \pm 1.0003}$ & $\mathbf{1.2984 \pm 0.6802}$ & $\mathbf{0.3482 \pm 0.1943}$ & -- \\
\bottomrule
\end{tabular*}
\end{table}

\noindent\textbf{Results on synthetic data.}
Table~\ref{tab:main_results} compares LongAgent with the three baseline methods, averaged across 360 synthetic datasets. LongAgent achieves the lowest RMSE ($1.7376$) and MAE ($1.2984 $), and the highest $R^2$ ($0.3482$) among all methods. Rule-based search is the second strongest method, which explores the same candidate space as LongAgent with a deterministic search policy. However, it follows the correlation ranking in search, while LongAgent leverages a rich history of evidence to guide the next move. Using two-sided Wilcoxon test, LongAgent outperforms rule-based search with statistical significance ($p < 0.05$).

\begin{table}[!ht]
\caption{Performance of LongAgent under three history configurations on 30 synthetic datasets. Values are mean $\pm$ standard deviation across datasets; ``+'' denotes evidence added cumulatively across rows, and $p$-values compare each augmented configuration with the basic history configuration.}
\label{tab:agent_variant_sensitivity}
\centering
\footnotesize
\begin{tabular*}{\textwidth}{@{\extracolsep{\fill}}lcccc@{}}
\toprule
History configuration & RMSE $\downarrow$ & MAE $\downarrow$ & $R^2$ $\uparrow$ & $p$ \\
\midrule
Basic history & $1.2982 \pm 0.4525$ & $0.9773 \pm 0.3184$ & $0.3993 \pm 0.2029$ & -- \\
\quad + Coverage summary & $1.2918 \pm 0.4558$ & $0.9730 \pm 0.3207$ & $0.4059 \pm 0.2043$ & 0.3223 \\
\quad + Transition evidence & $\mathbf{1.2841 \pm 0.4596}$ & $\mathbf{0.9670 \pm 0.3234}$ & $\mathbf{0.4145 \pm 0.2036}$ & 0.0488 \\
\bottomrule
\end{tabular*}
\end{table}

\noindent\textbf{Ablation study.}
Table~\ref{tab:agent_variant_sensitivity} investigates the effects of different evidence sources in the agent history. The basic history configuration contains search status and candidate history. By adding coverage summary to history, RMSE decreases from 1.2982 to 1.2918, and MAE decreases from 0.9773 to 0.9730, with $R^2$ increasing, all indicating better prediction performance. By further adding transition evidence to history, LongAgent achieves the lowest RMSE ($1.2841 \pm 0.4596$) and MAE ($0.9670 \pm 0.3234$), and the highest $R^2$ ($0.4145 \pm 0.2036$). Its RMSE is significantly lower than that obtained with basic history ($p < 0.05$), suggesting that transition evidence, such as past loss changes, provides informative signals for LongAgent to plan subsequent candidate updates.
\\

\begin{table}[!ht]
\caption{Comparison of LongAgent with baseline methods on a real clinical dataset, IDEA-FAST. Values report test performance on the subject-held-out split.}
\label{tab:ideafast_results}
\centering
\footnotesize
\begin{tabular*}{\textwidth}{@{\extracolsep{\fill}}lccc@{}}
\toprule
Method & RMSE $\downarrow$ & MAE $\downarrow$ & $R^2$ $\uparrow$ \\
\midrule
Greedy variable selection & 0.9957 & 0.8197 & 0.0173 \\
Random candidate search & 1.0012 & 0.8256 & 0.0063 \\
Rule-based search & 0.9939 & \textbf{0.8186} & 0.0208 \\
\textbf{LongAgent} & \textbf{0.9937} & 0.8187 & \textbf{0.0212} \\
\bottomrule
\end{tabular*}
\end{table}

\noindent\textbf{Results on IDEA-FAST.}
The real longitudinal data experiment compares LongAgent with the same three baseline methods on the IDEA-FAST dataset, with results reported in Table~\ref{tab:ideafast_results}. LongAgent achieves the lowest RMSE (0.9937) and highest $R^2$ (0.0212), outperforming the other methods. In terms of MAE, LongAgent only underperforms rule-based search by 0.0001. In terms of RMSE, LongAgent outperforms rule-based search by 0.0002, indicating comparable performance between the two methods on IDEA-FAST.

\section{Discussion and Conclusions}
We present a novel agent-based method, LongAgent, that can perform automated analysis for longitudinal medical data and discover predictive features for outcomes. LongAgent performs history-guided search over variable subsets, outcome-relative time windows, and aggregation functions. It constructs a rich history consisting of four sources of evidence to guide the search of the agent. Across 360 synthetic datasets, LongAgent achieved the lowest mean RMSE compared to three baseline methods. The ablation study demonstrates the usefulness of history guidance. On a real clinical dataset, the performance of LongAgent was slightly better or comparable to the strongest baseline. Current limitations include the use of a single shared time window and aggregation function for extracting features from selected variables. Future work will investigate variable-specific temporal representation learning and extend the method to additional longitudinal datasets with heterogeneous and missing data.

\subsubsection*{Acknowledgements}
The IDEA-FAST project has received funding from the Innovative Medicines Initiative 2 Joint Undertaking under grant agreement No. 853981. This Joint Undertaking receives support from the European Union's Horizon 2020 research and innovation programme and EFPIA and associated partners. This communication reflects the view of the authors and neither IMI nor the European Union and EFPIA are liable for any use that may be made of the information contained herein.
W.B. acknowledges the support of EPSRC CVD-Net Programme Grant (EP/Z531297/1) and BHF New Horizons Grant (NH/F/23/70013).

\bibliographystyle{splncs04}
\bibliography{references}

\end{document}